%% file: main.tex
\documentclass[runningheads]{llncs}

\usepackage{accv}

\usepackage{accvabbrv}

\usepackage{graphicx}
\usepackage{booktabs}

\usepackage{graphicx}

\usepackage{multirow}
\usepackage{algorithm}
\usepackage{algorithmic}
\usepackage{enumitem}

\usepackage{amssymb}
\usepackage{mathtools}

\usepackage[accsupp]{axessibility}  

\usepackage[pagebackref,breaklinks,colorlinks,citecolor=accvblue]{hyperref}

\usepackage{orcidlink}

\usepackage[textsize=tiny]{todonotes}

\begin{document}

\title{MT-SNN: Enhance Spiking Neural Network with Multiple Thresholds} 


\author{Xiaoting Wang\inst{1}, 
Yanxiang Zhang\inst{2}}


\authorrunning{Xiaoting Wang et al.}

\institute{Beijing University of Technology, China\\
\email{wxt22@bjut.edu.cn} \and
Individual, China\\
\email{stdcoutzyx@gmail.com}
}


\maketitle

\input{sections/abs}

\input{sections/intro}
\input{sections/related_works}

\input{sections/preliminary}

\input{sections/algorithm}

\input{sections/exp}

\input{sections/conclusions}

%
%
\bibliographystyle{splncs04}
\bibliography{main}
\end{document}

%% file: sections/abs.tex
\begin{abstract}
   Spiking neural networks (SNNs) present a promising energy efficient alternative to traditional Artificial Neural Networks (ANNs) due to their multiplication-free operations enabled by binarized intermediate activations. However, this binarization leads to precision loss, hindering the SNN performance.
   In this paper, we introduce Multiple Threshold (MT) approaches to significantly enhance SNN accuracy by mitigating precision loss. We propose two distinct modes for MT implementation, depending on the membrane update rule: parallel mode and cascade mode. MT-SNN models can be efficiently trained on standard hardwares like GPUs and TPUs, while retaining the multiplication-free advantage crucial for deployment on neuromorphic devices.  Our extensive experiments on CIFAR10, CIFAR100, ImageNet, and DVS-CIFAR10 datasets demonstrate that both MT modes substantially improve the performance of single-threshold SNNs, achieving higher accuracy with fewer time steps and comparable energy consumption. Moreover, MT-SNNs outperform state-of-the-art (SOTA) results. Notably, with MT, a Parametric-Leaky-Integrate-Fire (PLIF) based ResNet-34 architecture reaches 72.17\% accuracy on ImageNet with a single time step, surpassing the previous SOTA by 2.75\% despite using 4 steps.


\keywords{Spiking Neural Network \and Image Classification \and Computer Vision }
\end{abstract}

%% file: sections/intro.tex
\section{Introduction}

Recent advances in Artificial Neural Networks (ANNs), particularly deep learning, have revolutionized fields such as computer vision\cite{he2016deep, tolstikhin2021mlp}, natural language processing\cite{brown2020language, devlin2018bert, vaswani2017attention}, and multi-modal learning\cite{lu2019vilbert,radford2021learning}. However, ANNs often demand substantial computational resources. A standard convolutional neural network for ImageNet-1k classification, for instance, consumes approximately 250 watts\cite{roy2019towards}, starkly contrasting with the 20 watts utilized by the human brain to simultaneously process diverse tasks\cite{laughlin2003communication}.


Spiking Neural Networks (SNNs), inspired by the brain's biological neurons\cite{maass1997networks}, are emerging as a next-generation neural model. SNNs have garnered increasing interest due to their unique capabilities, including temporal information processing, energy efficiency\cite{roy2019towards}, and high biological plausibility\cite{gerstner2014neuronal}. By leveraging binary spike signals, SNNs achieve low energy consumption through multiplication-free inference and by bypassing computations on zero-valued inputs or activations\cite{roy2019towards}. Neuromorphic hardware like TrueNorth \cite{akopyan2015truenorth}, Lohi \cite{davies2018loihi} and Tianjic \cite{pei2019towards} have demonstrated the potential of SNNs to achieve energy savings of orders of magnitude.


\begin{figure*}
    \centering
    \includegraphics[width=12cm, height=6.4cm]{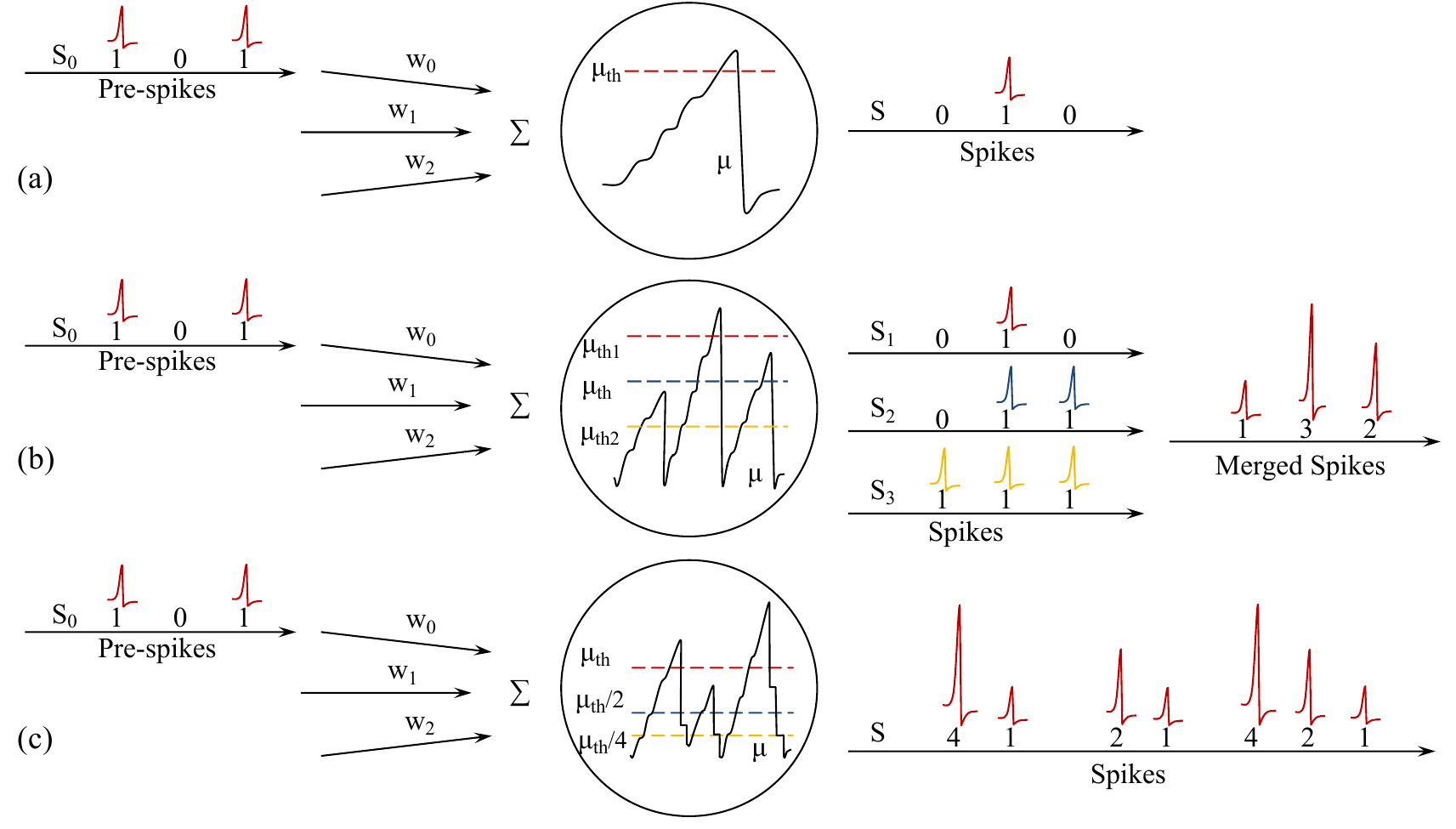}
    \caption{Spikes with single threshold and multiple threshold, $\mu$ is membrane and $S$ is output spike, $th$ is short for threshold.
    (a): Single Threshold (ST): Spikes are fired only if the membrane exceeds $\mu_{th}$.
    (b): PARALLEL MT: Three independent thresholds are applied, thus three spike sequences are generated and merged.
    (c): CASCADE MT: Three thresholds are applied and will be checked sequentially from highest to lowest, the fired spikes are weighted before aggregation.
    }
    \label{fig:st_mt}
\end{figure*}

Despite their inherent energy efficiency and biological plausibility, SNNs face several challenges that impede their broader adoption in mainstream machine learning.  A primary obstacle is the inevitable information loss that occurs when converting floating-point activations into binary spikes. Although increasing the number of time steps can partially alleviate this, performance remains limited when time steps are constrained. Additionally, many popular tasks lack intrinsic temporal structure, necessitating inefficient preprocessing to adapt them for SNNs. For instance, in static image classification task, images need to be encoded with a time dimension, requiring multiple forward passes, while ANNs can process them with only one pass. Finally, SNNs are hindered by both software and hardware limitations. Popular machine learning frameworks such as Pytorch \cite{paszke2019pytorch}, Jax \cite{bradbury2018jax} and Tensorflow \cite{abadi2016tensorflow} lack efficient, generalized support for spike-based computations. Moreover, standard hardware like GPUs and TPUs, optimized for floating-point arithmetic, are not ideal for SNNs' multiplication-free operations.

In this work, we introduce the Multiple Threshold (MT) approach to address the precision loss inherent in SNNs. MT also mitigates the limitations imposed by the temporal structure of data, enabling SNNs to achieve high accuracy with fewer time steps. We propose two distinct MT modes, depending on the membrane update rule:


\begin{itemize}[itemsep=2pt,topsep=0pt,parsep=0pt]
    \item PARALLEL MT: Multiple thresholds are applied independently to determine spike firing. with the results aggregated across thresholds to enhance precision. See \cref{fig:st_mt}-(a, b) for a visual comparison with single threshold SNNs.
    \item CASCADE MT: A series of proportionally decreasing thresholds are employed to sequentially trigger spike firing, followed by a weighted sum of the spiking results. See \cref{fig:st_mt}-(a, c) for a visual comparison with single threshold SNNs.
\end{itemize}

The MT algorithm demonstrably enhances SNN performance, achieving higher accuracy with reduced time steps and comparable energy consumption. This substantially eases the computational burden of SNN training on existing hardware. Furthermore, by strategically adding branches in the layer subsequent to the LIF neuron, the trained model remains compatible with neuromorphic hardwares, fully retaining the inherent advantages of spike-based computation.



Our contributions can be summarized as follows:

\begin{itemize}[itemsep=2pt,topsep=0pt,parsep=0pt]
    \item We introduce the MT algorithm, a versatile, modular approach that strategically utilizes multiple thresholds in either parallel or cascade configurations to enhance the precision of SNNs across diverse architectures.
    \item We demonstrate the effectiveness and robustness of MT across various SNN architectures (VGG, ResNet) and datasets (CIFAR-10, CIFAR-100, ImageNet, DVS-CIFAR10), showcasing its broad applicability.
    \item Our MT-SNNs significantly elevate the performance ceiling of SNNs, achieving results that surpass those of single-threshold models, even with substantial increases in time steps. Furthermore, MT-SNNs outperform the majority of existing approaches, match current state-of-the-art (SOTA) results, and even surpass SOTA performance with much fewer steps and energy.
\end{itemize}

%% file: sections/related_works.tex
\section{Related Works}

\subsection{Learning Algorithms}

After the emergence of SNNs, how to train the models effectively stays as an active area. The earliest learning algorithms were based on biological plausible local learning rules, like Hebbian learning \cite{hebb2005organization} and Spike-Timing-Dependent Plasticity \cite{bi1998synaptic}.
However, such methods were only suitable for shallow SNNs, and the performance is far below methods mentioned below.

ANN-To-SNN methods are capable of obtaining high-quality SNNs by leveraging the knowledge from well-trained ANNs and converting the ReLU activations to spike activations based on rate coding \cite{sengupta2019going, deng2021optimal}. However, achieving near lossless conversion requires a considerable amount of time steps ($>200$) to accumulate the spikes, which may significantly increase the latency.

Backpropagation with surrogate gradient is widely used in recent researches to achieve high-quality deep models  \cite{lee2016training,wu2018spatio,lee2020enabling}. The utilization of surrogate gradient function enables end-to-end backpropagation in SNN training, thus training knowledge of ANNs can be transferred to SNNs. The remarkable robustness of the surrogate gradient based algorithms were studied 
in \cite{zenke2021remarkable, neftci2019surrogate}, demonstrating that SNNs trained with surrogate gradient can achieve competitive performance with ANNs. Backpropagation with surrogate functions requires much fewer steps compared to ANN-To-SNN methods as it's not dependent of rate coding \cite{YujieWu2019DirectTF}.

\subsection{Model architecture}

Backpropagation with surrogate gradient unlocks the model architecture migration from ANNs to SNNs. Yujie et al. \cite{YujieWu2019DirectTF} initially applied direct training of various sized deep convolutional neural networks on both static and neuromorphic datasets with normalization and rate coding. Hanle et al. \cite{HanleZheng2020GoingDW} explored ResNet with time-dependent batch normalization and validated the architecture on ImageNet for the first time.

Spike-element-wise ResNet \cite{WeiFang2021DeepRL} and membrane based ResNet \cite{hu2021advancing} pushed the depth of the architecture to more than 100 layers. Attention machanism was introduced into SNN by Man Yao et al. \cite{yao2022attention}, achieving comparable and even better performance compared to the ANN counterpart on large-scale datasets. Furthermore, Spiking Transformer was proposed by combining self-attention with spiking neural network \cite{zhou2022spikformer} . There will be more works on such model migrations in the foreseeable future.

\subsection{Information Loss}

Information loss in SNNs attracts increasing attentions recently. Soft Reset was leveraged to reserve the information in the residual potential, and Membrane Potential Rectifier was proposed to alleviate the precision loss in quantization \cite{guo2022reducing}. Yufei Guo et al. \cite{guo2022real} leveraged a re-parameterization technique to decouple training and inference stage, and enhance the presentation capacity b learning real-valued spikes during training and transferring the capacity to model parameters by not sharing convolution kernels. The MT approaches in this paper fall into the same domain with Yufei Guo et al. \cite{guo2022reducing, guo2022real}, which provides another option to reduce information loss in spike activations.

\subsection{Quantization}

Quantization is also a very promising approach to reduce computational cost and memory footprint while maintaining competitive performance \cite{rastegari2016xnor, bulat2019xnor}. Spiking could be considered as the 1-bit quantization, thus SNN training could be considered as the activation quantization aware ANN training. 

The CASCADE MT approach proposed in this paper is a general way to expand the 1-bit quantization property of SNNs into multiple bits, saving the precision loss.

%% file: sections/preliminary.tex
\section{Preliminary}

\subsection{Parametric Leaky Integrate and Fire Model}

Parametric Leaky Integrate and Fire (PLIF) neuron model \cite{WeiFang2020IncorporatingLM} is adopted as the basic unit in the model architectures in this paper. PLIF includes the following discrete time equations:

\begin{equation}
\label{eqn:neuron_dynamics}
    H[t] = f(V[t-1], X[t]),
\end{equation}
\begin{equation}
\label{eqn:neuron_spike}
    S[t] = \Theta(H[t] - V_{th}),
\end{equation}
\begin{equation}
\label{eqn:neuron_update}
    V[t] = H[t](1 - S[t]) + V_{reset}S[t]
\end{equation}

Where $X[t]$ is the input current at time step t from the previous layer, $H[t]$ and $V[t]$ represent the membrane potential after neuronal dynamics and spike triggering respectively. $V_{th}$ is the firing threshold. $\Theta(x)$ is the Heaviside step function.



$V_{reset}$ denotes the reset potential. The function in \cref{eqn:neuron_dynamics} describes the neuronal dynamics, which can represent different model with different forms. \cref{eqn:lif} describes the dynamics of the Leaky Integrate-and-Fire(LIF).



\begin{equation}
\label{eqn:lif}
    H[t] = V[t-1] + \frac{1}{\tau}(X[t] - (V[t-1] - V_{reset}))
\end{equation}

In \cref{eqn:lif}, the $\tau$ is the membrane time constant, which is made learnable with \cref{eqn:learnable_plif} \cite{WeiFang2020IncorporatingLM}.

\begin{equation}
\label{eqn:learnable_plif}
    \tau = 1 + exp(-a) \in (1, +\infty)
\end{equation}

Where $a$ is a learnable variable.

Furthermore, when $V_{reset} = 0$, the PLIF neuron becomes \cref{eqn:plif_lstm}.

\begin{equation}
\label{eqn:plif_lstm}
    H[t] = (1 - \frac{1}{\tau})V_{t-1} + \frac{1}{\tau}X_t
\end{equation}

which is analogous to the input gate and forget gate in Long Short-Term Memory(LSTM) networks \cite{SeppHochreiter1997LongSM}. In this paper, We set $V_{reset}=0$ and implement the PLIF model with formula in \cref{eqn:plif_lstm}.

\subsection{Surrogate Function}

The surrogate function used in this paper is the popular rectangular function proposed by Yujie Wu et al. \cite{YujieWu2019DirectTF}, given by \cref{eqn:surrogate_func} below.

\begin{equation}
\label{eqn:surrogate_func}
    \frac{\partial S[t]}{\partial H[t]} = \frac{1}{a}sign(|H[t] - V_{th}| \leq \frac{a}{2})
\end{equation}

Where a is a hyper-parameter and usually set to 1.







%% file: sections/algorithm.tex

\section{Multi-Threshold SNN}

Increasing the number of time steps often improves SNN performance, as seen in Fig. \ref{fig:st_mt_cifar10_cifar100}.  Single-Threshold (ST) SNNs typically underperform their ANN counterparts at lower time steps, but accuracy improves with increasing time steps. For VGG-like architectures, SNNs can match or even surpass ANN accuracy when the number of time steps exceeds 3.



This observation suggests that the precision lost in converting floating-point activations to spikes in SNNs can be compensated by using multiple time steps.  In this paper, we propose an alternative solution: the Multiple Threshold (MT) approach, which aims to preserve precision within activations even with a single time step. In the following sections, we detail the parallel and cascade modes of MT implementation.


\subsection{PARALLEL MT}
\label{sec:parallel_mt}

The idea is illustrated in \cref{fig:st_mt}-(a, b). In \cref{fig:st_mt}-(a), a single threshold $\mu_{th}$,  discretizes the input membrane potential into binary values 0 or 1, leading to a loss of information regarding how much the potential exceeds or falls below the threshold. In \cref{fig:st_mt}-(b), we introduce two auxiliary thresholds $\mu_{th1,th2}$. These thresholds generate two additional spike sequences based on their respective comparisons with the membrane potential. Summing these three spike sequences results in a more accurate representation of the original, continuous membrane potential.


To formulate the approach, a series of parameters are introduced in, namely $[\Delta_1, ..., \Delta_n]$, and integrated into \cref{eqn:neuron_spike}:

\begin{equation}
\label{eqn:neuron_multiple_spikes}
    S[t]_{i} = \Theta(H[t] - V_{th} - \Delta_i), i \in [1, 2, ..., n]
\end{equation}

And \cref{eqn:neuron_update} is still used to update the membrane. After obtaining all spikes for each threshold, the final outputs are obtained by summing the spikes together.

\begin{equation}
\label{eqn:neuron_multiple_sum_spikes}
    S[t]_{sum} = S[t] + \Sigma{S[t]_{i}}
\end{equation}

However, after applying multiple thresholds, from \cref{eqn:neuron_multiple_sum_spikes}, the outputs have other possible values other than 0/1, which breaks the multiplication-free property. To preserve the property, the subsequent convolution layer can be involved in to perform an equivalent operation in \cref{eqn:neuron_sum}, right side remains multiplication-free.

\begin{equation}
\label{eqn:neuron_sum}
    Conv(S[t]_{sum}) = Conv(S[t]) + \Sigma{Conv(S[t]_{i})}
\end{equation}

This formulation allows the left-hand side of Eq. \ref{eqn:neuron_sum} to be deployed on standard hardware like GPUs or TPUs, while the right-hand side remains compatible with neuromorphic hardware.


\subsection{CASCADE MT}
\label{sec:cascade_mt}

The concept of CASCADE MT is illustrated in \cref{fig:st_mt}-(a, c). As shown in \cref{fig:st_mt}-(c), we employ three sequentially decreasing thresholds ($\mu_{th}$, $\mu_{th}/2$, $\mu_{th}/4$), checked from highest to lowest, A soft reset mechanism allows the remaining membrane potential to be utilized by lower thresholds, maximizing information retention.

To formulate this approach, a new parameter $b$ is introduced to determine the number of threshold reductions. By default, the threshold reduction factor is set to be 2, creating a close analogy to activation quantization.



The forward path of CASCADE MT is outlined in \cref{alg:cascade_mt}. For a given membrane potential $H[t]$, firing is performed b times, in each iteration, the threshold is halved, the membrane potential is the residual from the previous step and the resulted spikes will be weighted by $2^{b-1-i}$.


\begin{figure}[ht]
    \centering
    \vspace{-1.2cm}
    \begin{minipage}{.60\linewidth}
        \begin{algorithm}[H]
            \caption{CASCADE Multiple Threshold}
            \label{alg:cascade_mt}
            \begin{algorithmic}
                \ENSURE{$H[t]$}\COMMENT{Membrane at time t}
                \ENSURE{$V_{th}$}\COMMENT{Initial Threshold}
                \ENSURE{$b$}\COMMENT{Number of bits}
                \REQUIRE{$S[t]_{sum}$}\COMMENT{The sum of output spikes}
                \STATE $h\gets{H[t]}$
                \STATE $S[t]_{sum}\gets{0}$
                \FOR{$i$\ in $\left[ 0 ... b\right)$ }
                    \STATE $S[t]_i\gets{\Theta(h - V_{th} / 2^i)}$
                    \STATE $h\gets{h-S[t]_i * V_{th} / 2 ^ i}$
                    \STATE $S[t]_{sum}\gets{S[t]_{sum} * 2 + S[t]_i}$
                \ENDFOR
            \end{algorithmic}
        \end{algorithm}
    \end{minipage}
    \vspace{-0.6cm}
\end{figure}

Similar to PARALLEL MT, the resulting $S[t]_{sum}$ in \cref{alg:cascade_mt} is not strictly 0/1 spikes. To maintain compatibility with neuromorphic hardware, we involve the weights in the next convolutional layer to perform an equivalent operation, as shown in the right-hand side of \cref{eqn:cascade_neuron_sum}. 
While the weights of convolution layers in right side \cref{eqn:cascade_neuron_sum} are duplicated by multiplying $2^i$.


\begin{equation}
\begin{split}
\label{eqn:cascade_neuron_sum}
    Conv(S[t]_{sum}) & = \Sigma{ Conv_{W_{i}}(S[t]_{i}) }\\
    W_{i} & = W * 2^i, i \in [0, b)
\end{split}
\end{equation}


%% file: sections/exp.tex
\section{Experiments}

We evaluate the proposed MT methods on static datasets CIFAR10\cite{krizhevsky2009learning}, CIFAR100\cite{krizhevsky2009learning}, ImageNet\cite{krizhevsky2012imagenet} and one neuromorphic dataset CIFAR10-DVS\cite{li2017cifar10}. Various model architectures such as VGG, ResNet-19/34 are implemented to demonstrate the versatility of the methods. All code is implemented with Tensorflow, and experiments are conducted on TPU.

\subsection{Experiment Settings}

\subsubsection{Network Architectures}

\begin{table}[tb]
    \caption{Model Configurations}
    \label{tab:model_configs}
    \centering
    \begin{tabular}{ccccccccc}
        \toprule
        Model & Dataset  & $N_{stages}$ & $N_{conv}$ & Conv Filters & $N_{fc}$ & FC-Hidden & $N_{params}$ \\
        \midrule
        VGG-8 & CIFAR10(100) & 2 & 3,3 & 256,256 & 2 & 2048 & 36.72M \\
        VGG-9 & CIFAR10(100) & 3 & 2,2,3 & 256,512,512 & 2 & 1024 & 19.72M \\
        ResNet19 & CIFAR10(100) & 3 & 6,6,4 & 128,256,512 & 1 & N/A & 14.04M \\
        ResNet34 & ImageNet & 4 & 6,8,12,4 & 128,256,512,1024 & 1 & N/A & 277.5M \\
        VGG-12 & DVS-CIFAR10 & 4 & 2,2,3,3,3 & 128 & 2 & 512 & 2.88M \\
        \bottomrule
    \end{tabular}
\end{table}

\begin{table}[!htb]
    \caption{Performance comparison of MT with SoTA methods on CIFAR datasets}
    \label{tab:cifar10_cifar100_dvscifar10_results}
    \centering
    \begin{tabular}{ccccc}
        \toprule
        Dataset & Name & Model & Steps & Accuracy(\%) \\
        \midrule
        \multirow{17}{*}{CIFAR10/100} & \multirow{3}{*}{STBP-tdBN\cite{HanleZheng2020GoingDW}} & \multirow{3}{*}{ResNet-19} & 2 & 92.34 / - \\
        & & & 4 & 92.92 / - \\
        & & & 6 & 93.16 / -\\
        \cline{2-5}
        & PLIF\cite{WeiFang2020IncorporatingLM} & VGG-8 & 8 & 93.50 / - \\
        \cline{2-5}
        & \multirow{3}{*}{InfLoR-SNN\cite{guo2022reducing}} & \multirow{2}{*}{ResNet-19} & 2 & 94.44 / 75.56 \\
        & & & 4 & 96.27 / 78.42 \\
        & & VGG-16 & 5 & 94.06 / 71.56 \\
        \cline{2-5}
        & \multirow{2}{*}{Real Spike\cite{guo2022real}} & \multirow{2}{*}{ResNet-19} & 2 & 94.01 / - \\
        & & & 4 & 95.60 / - \\
        \cline{2-5}
        & \multirow{3}{*}{DSpike\cite{YuhangLi2021DifferentiableSR}} & \multirow{3}{*}{ResNet-18} & 2 & 93.13 / 71.68 \\
        & & & 4 & 93.66 / 73.35 \\
        & & & 6 & 94.25 / 74.24 \\
        \cline{2-5}
        & \multirow{5}{*}{Ours} & \multirow{5}{*}{VGG-9} & 1 & 95.17 / 74.80 \\
        & & & 2 & 95.08 / 74.08 \\
        & & & 3 & 95.25 / 75.04 \\
        & & & 4 & \textbf{95.34} / 75.92 \\
        & & & 5 & 95.14 / \textbf{76.59} \\
        \cline{2-5}
        & \multirow{5}{*}{Ours} & \multirow{5}{*}{ResNet-19} & 1 & 95.31 / 76.92 \\
        & & & 2 & 95.33 / 77.40 \\
        & & & 3 & 95.68 / 77.13 \\
        & & & 4 & 95.64 / \textbf{78.07} \\
        & & & 5 & \textbf{95.79} / 77.33 \\
        \hline
        \multirow{6}{*}{DVS-CIFAR10} & STDP-tdBN\cite{HanleZheng2020GoingDW} & ResNet-17/19 & 10 & 67.80 \\
        & PLIF\cite{WeiFang2020IncorporatingLM} & Conv+FC & 20 & 74.80 \\
        & SEW-ResNet\cite{WeiFang2021DeepRL} & SEW-ResNet & 16 & 74.40 \\
        & DSpike\cite{YuhangLi2021DifferentiableSR} & ResNet-18 & 10 & 75.40 \\
        & MS-ResNet\cite{hu2021advancing} & ResNet-20 & - & 75.56 \\
        & InfLoR-SNN\cite{guo2022reducing} & ResNet-19 & 10 & 75.50 \\
        \cline{2-5}
        & \multirow{2}{*}{Ours} & ST-VGG-12 & 5 & 75.90 \\
        & & MT-VGG-12 & 5 & \textbf{76.30} \\
        \bottomrule
    \end{tabular}
\end{table}

The model configurations are detailed in \cref{tab:model_configs}. Convolutional layers with the same feature map size are considered as a stage, sharing the same number of filters. VGG models terminate with two fully connected layers (hidden and output), while the ResNet models have only the output layer. 

VGG-8, proposed by Wei et al.\cite{WeiFang2020IncorporatingLM}, is a compact architecture for CIFAR10. Due to the large output size of the last pooling layer, 92\% of its parameters reside in the first dense layer. We introduce VGG-9 to deepen the network and reduce model size.


Each convolution layer is followed by a batch normalization layer and a PLIF layer to fire spikes, the first Conv-BN-PLIF serves as the encoder, converting image pixels into spikes.

Following Hu et al.\cite{hu2021advancing}, residual connections in ResNet models operate at the membrane potential level. After convolutional layers, a pooling layer reduces the output feature map to 1D before the final output layer. All convolutional layers use 3x3 kernels, except for the first layer in ResNet-34 for ImageNet, which uses a 7x7 kernel.


\subsubsection{Datasets}

The CIFAR-10(100) datasets datasets contain 50,000 images for training and 10,000 images for validation with 32x32 pixels in 10(100) classes. Random horizontal flip, random rotation by 15 degree and random height, width shift by 10\% fraction are used as the data augmentation approaches.

The CIFAR10-DVS dataset is the neuromorphic version of the CIFAR-10 dataset, it's split into 9,000 images for training and 1,000 images for validation\cite{YujieWu2019DirectTF}. Events are converted into frames by evenly accumulating into T slices\cite{WeiFang2020IncorporatingLM}. No data augmentation is used.

The Imagenet dataset is a large-scale dataset with 1,281,167 images for training and 50,000 images for validation. The size of model inputs is $224\times224$. For training, images are first resized by setting the shorter edge to a random value within the range $[256, 384]$, followed by a random crop of $224\times224$ patches. After cropping, randomly flipping vertically and horizontally are leveraged as data augmentation. For validation, images are resized by setting the shorter edge to 320 and then center-cropped to $224\times224$. 

\subsubsection{Training Setup}

We employ the SGD optimizer with momentum set to 0.9 in all experiments. The learning rate follows a cosine decay schedule\cite{loshchilov2016sgdr}, decreasing to zero throughout training. For CIFAR10, CIFAR100, and CIFAR10-DVS, we use a batch size of 100 and an initial learning rate of 0.01. The learning rate is reset and decays to zero every 200 epochs with a full restart, and training proceeds for a total of 600 epochs. For ImageNet, we increase the batch size to 5,000 to expedite training, which lasts for 100 epochs. To stabilize training, we warm up the learning rate linearly from 0 to 0.1 during the first 10 epochs, then cosine decay it to 0.001 over the next 80 epochs, and hold it at 0.001 for the final 10 epochs. No learning rate restarts are used for ImageNet.


\subsubsection{Spiking Thresholds}

We use a spiking threshold of 1.0 for the single-threshold baseline. For PARALLEL MT, we maintain the threshold at 1.0 but introduce additional $\Delta = [-0.3, 0.3]$. As discussed in \cref{sec:parallel_mt}, this is equivalent to using three thresholds: [0.7, 1.0, 1.3]. For CASCADE MT, we set $b=4$, resulting in 4 thresholds as described in Section \ref{sec:cascade_mt}. Through experimentation, we found that an initial spiking threshold of 4 performs best for ImageNet, while 2 is optimal for CIFAR datasets. We include ablation studies to explore the effects of reducing the number of $\Delta$ for PARALLEL MT and the number of bits for CASCADE MT.


\subsection{Results}

\subsubsection{CIFAR10 / CIFAR100}

\begin{figure}[ht]
    \vskip 0.2in
    \begin{center}
    \includegraphics[width=12cm, height=6.5cm]{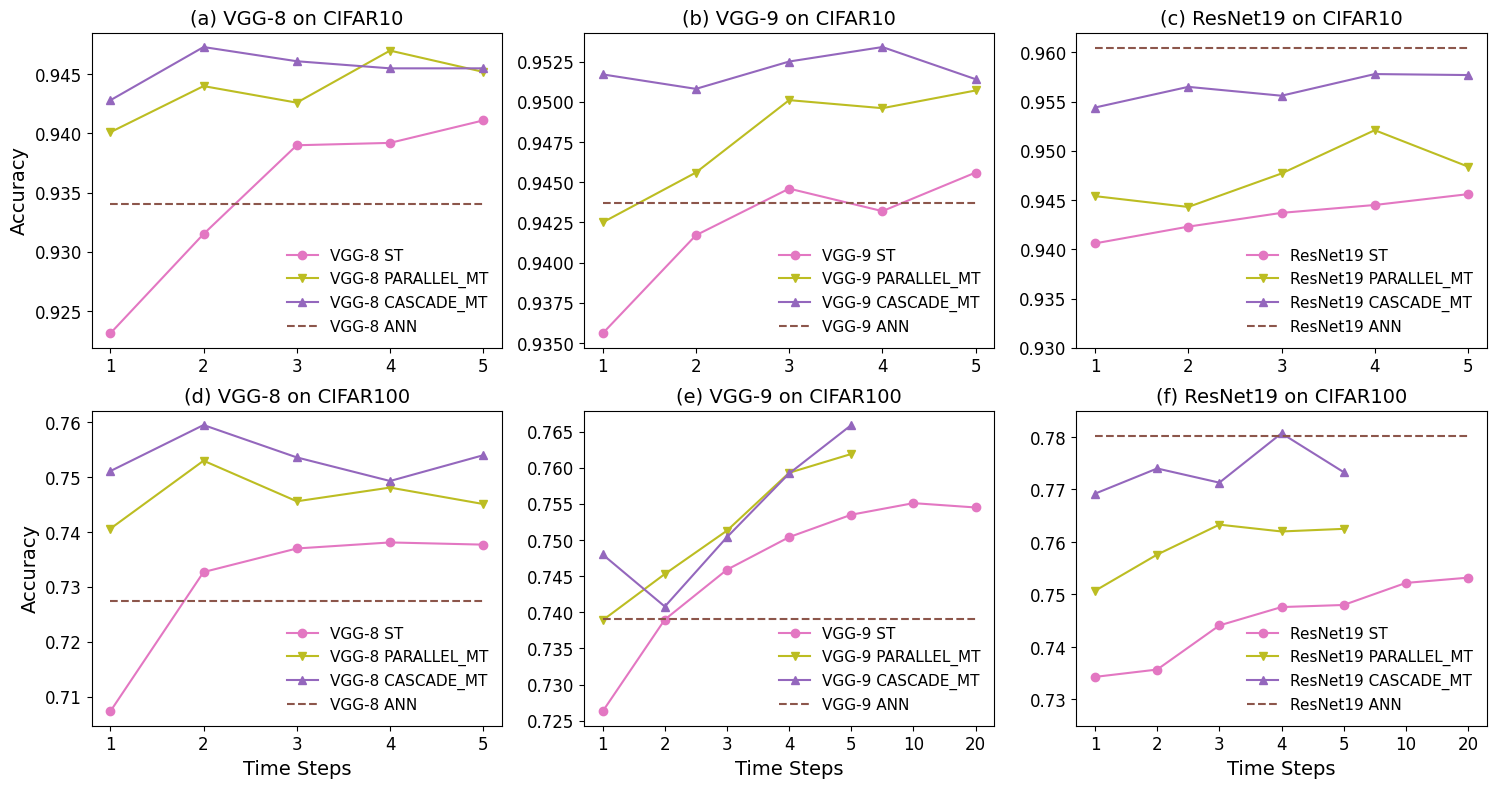}
    \vspace*{-3mm}
    \caption{Performance comparison of ST and MT on CIFAR10 / CIFAR100 with various steps. There are four arms in each figure, namely ANN, SNN with Single Threshold(ST), SNN with PARALLEL MT, CASCADE MT.}
    \label{fig:st_mt_cifar10_cifar100}
    \end{center}
    \vskip -0.2in
\end{figure}

\cref{fig:st_mt_cifar10_cifar100} demonstrates the accuracy advantages of PARALLEL MT and CASCADE MT SNNs over both ST(Single Threshold) SNNs and their ANN counterparts on CIFAR10 and CIFAR100 datasets. Across all VGG-8, VGG-9, and ResNet-19 models, and for time steps ranging from 1 to 5, MT-SNNs consistently outperform their ST counterparts. And CASCADE MT generally yields the best results.

Notably, MT-enabled VGGs surpass its ANN counterpart at a single time step, while Spiking ResNet-19 achieves comparable accuracy to the ANN at $step=4$. MT significantly raises the performance ceiling for SNNs. Even with extensive time steps (10 and 20), ST-SNNs remain inferior to MT approaches, as shown in \cref{fig:st_mt_cifar10_cifar100}-(e,f).

\cref{tab:cifar10_cifar100_dvscifar10_results} compares our results with previous state-of-the-art (SOTA) models on similar architectures. Our MT ResNet-19 achieves remarkable accuracies of 95.79\% on CIFAR10 and 78.07\% on CIFAR100, nearly matching the previous SoTA InfLoR-SNN \cite{guo2022reducing}. Impressively, our ResNet-19 model with just one step could reach 95.31\% on CIFAR10 and 76.92\% on CIFAR100, exceeding the InfLoR-SNN at $step=2$ and outperforming the other prior works.

\subsubsection{ImageNet}

\cref{tab:imagenet_results} summarizes the our results and comparisons with previous SOTA models on ImageNet. Our ST SNN implementation reaches 66.79\% accuracy, on par with SEW-ResNet\cite{WeiFang2021DeepRL}, InfLoR-SNN\cite{guo2022reducing} and Real Spike\cite{guo2022real} with $step=4$. Applying MT leads to signficant improvements, with CASCADE MT achieving 72.17\% accuracy at just one time step, surpassing the previous SOTA MS-ResNet\cite{hu2021advancing} even at $step=6$.


\begin{table}[tb]
    \caption{ImageNet Performance comparison of our methods with SOTA on ResNet-34}
    \label{tab:imagenet_results}
    \centering
    \begin{tabular}{ccc}
        \toprule
        Name & Steps & Accuracy(\%) \\
        \midrule
        SEW-ResNet\cite{WeiFang2021DeepRL} & 4 & 67.04 \\
        DSpike\cite{YuhangLi2021DifferentiableSR} & 6 & 68.19 \\
        MS-ResNet\cite{hu2021advancing}  & 6 & 69.42 \\
        InfLoR-SNN\cite{guo2022reducing}  & 4 & 65.54 \\
        Real Spike\cite{guo2022real}  & 4 & 67.69 \\
        \hline
        ST &  4 & 66.79 \\
        PARALLEL-MT & 1 & 69.32 \\
        CASCADE-MT & 1 & \textbf{72.17} \\
        \bottomrule
    \end{tabular}
\end{table}

\subsubsection{CIFAR10-DVS}

\begin{figure}[tb]
    \vskip 0.2in
    \begin{center}
    \includegraphics[width=4.8cm, height=3.8cm]{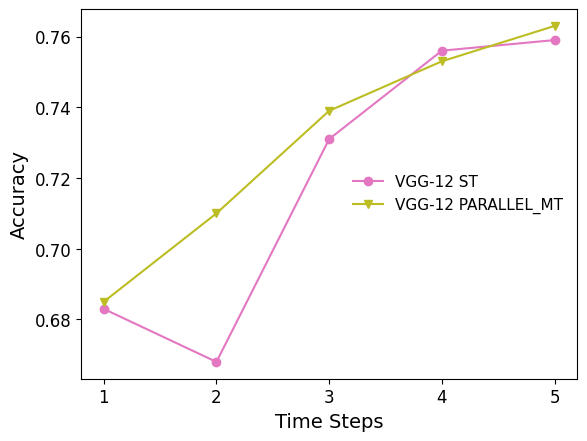}
    \vspace*{-3mm}
    \caption{Performance comparison of ST and PARALLEL MT on CIFAR10-DVS with various steps.}
    \label{fig:vgg12_dvscifar10_results}
    \end{center}
    \vskip -0.2in
\end{figure}

We evaluate MT on the CIFAR10-DVS dataset using the VGG-12 model in \cref{tab:model_configs}. \cref{fig:vgg12_dvscifar10_results} demonstrates significant accuracy gains with MT, particularly at early time steps (less than 4). Our model ultimately achieves 76.3\% accuracy, surpassing previous models with similar architectures in \cref{tab:cifar10_cifar100_dvscifar10_results}.

However, the performance gap between ST and MT narrows with increasing time steps. We also observe substantial accuracy fluctuations on CIFAR10-DVS, likely due to its limited size. Evaluating MT on larger DVS datasets is an important direction for future work, although such datasets are currently scarce.



\subsection{Ablation Study}

Extensive ablation studies are conducted to pinpoint the factors driving performance improvements.

\begin{figure*}[tb]
    \vskip 0.2in
    \begin{center}
    \includegraphics[width=12cm, height=3.6cm]{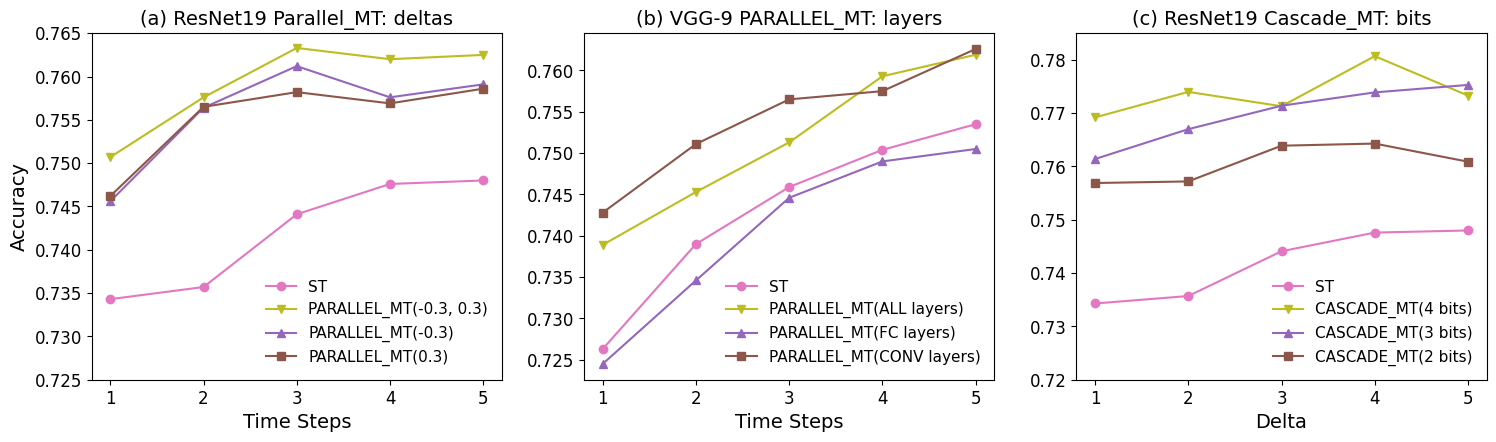}
    \vspace*{-3mm}
    \caption{Ablation study of PARALLEL MT and CASCADE MT on CIFAR100. (a): Impact of thresholds $\Delta$ in PARALLEL MT. (b): Impact of layer placement in PARALLEL MT. (c): Impact of bit number in CASCADE MT.}
    \label{fig:ablation_exp}
    \end{center}
    \vskip -0.2in
\end{figure*}

First, we analyze PARALLEL MT settings. Recall that our previous experiments used $\Delta=[-0.3, 0.3]$ for all LIF layers. \cref{fig:ablation_exp}-(a) shows that using $\Delta=[-0.3]$ and $\Delta=[0.3]$ individually yields similar improvements, roughly 80\% of the combined gain. Additionally, we find that applying PARALLEL MT to fully-connected layers is detrimental in \cref{fig:ablation_exp}-(b). Limiting MT to convolutional layers, particularly for VGG-9 with 42\% of parameters in fully-connected layers, yields superior results.


Second, we explore the impact of the number of bits (b) in CASCADE MT. \cref{fig:ablation_exp}-(c) demonstrates that increasing b improves accuracy, but with diminishing returns. Notably, $b=2$ achieves roughly 50\% of the total improvement from ST to CASCADE MT ($b=4$), while $b=3$ yields near-optimal accuracy.


\subsection{Sparsity, Efficiency and Energy Cost}

We measure sparsity, computation, and energy consumption following Li et al.\cite{YuhangLi2021DifferentiableSR} and Rathi et al.\cite{rathi2020diet}. We use ${s}\times{T}\times{A}$ to estimate additions in SNNs, where $s$ is average sparsity, $T$ is the time step, and $A$ is the total additions in the corresponding ANN. Energy is based on 45nm CMOS technology, with Multiply-ACcumulate (MAC) costing $4.6pJ$ and accumulation costing $0.9pJ$.

\begin{figure*}[tb]
    \vskip 0.2in
    \begin{center}
    \includegraphics[width=12cm, height=4.0cm]{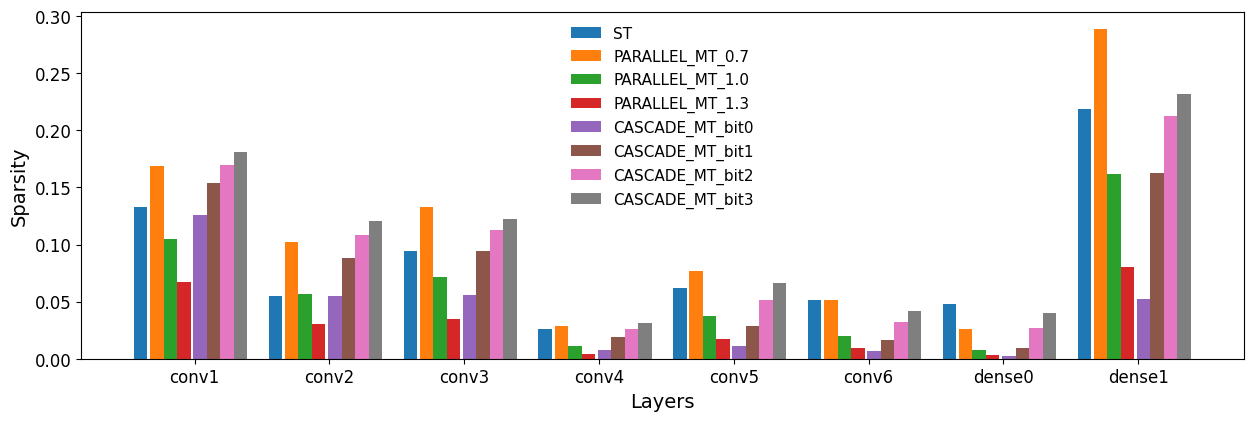}
    \vspace*{-3mm}
    \caption{Activation Sparsity of VGG-9 layers on CIFAR100}
    \label{fig:activation_sparsity}
    \end{center}
    \vskip -0.2in
\end{figure*}

\cref{fig:activation_sparsity} shows input sparsity for each VGG-9 SNN layer (excluding the first layer, conv0, which takes raw images). ST, PARALLEL MT with two additional thresholds, and CASCADE MT with 4 bits are compared. ST models exhibit low sparsity (2.6\% to 13.3\%) across most layers, except for 21.9\% in the final dense layer. This low sparsity is key to energy efficiency. While MT introduces multiple activations, average sparsity remains close to ST levels. Thus, additions increase roughly linearly with the number of thresholds (PARALLEL MT) or bits (CASCADE MT). In this example, PARALLEL MT has about 3x and CASCADE MT has 4x the additions of ST.


\cref{tab:cifar100_energy} presents operations and energy consumption for VGG-9 models on CIFAR100. PARALLEL MT at $step=5$ slightly outperforms ST at $step=15$, while CASCADE MT at $step=5$ significantly outperforms ST at $step=20$, improving accuracy by 1.10\% with less energy. Notably, PARALLEL MT efficiency can be further improved by removing MT from fully-connected layers, and CASCADE MT benefits from reducing bits from 4 to 3, as per our ablation study.


\begin{table}[tb]
    \caption{Engergy comparison of different VGG-9 model variant on CIFAR100}
    \label{tab:cifar100_energy}
    \centering
    \begin{tabular}{cccccc}
        \toprule
        Model & Step & Accuracy(\%) & \#MUL & \#ADD & Energy \\
        \midrule
        ANN & - & 73.54 & 1978M & 1978M & 10.8J \\
        \hline
        \multirow{5}{*}{ST} & 1 & 72.63 & 7.07M & 182.2M & 0.19J \\
        & 3 & 74.59 & 21.23M & 546.61M & 0.59J \\
        & 4 & 75.04 & 28.31M & 782.82M & 0.83J \\
        & 15 & 75.46 & 106.17M & 2733.06M & 2.96J \\
        & 20 & 75.45 & 141.56M & 3644.08M & 3.93J \\
        \hline
        \multirow{2}{*}{PARALLEL MT} & 1 & 73.89 & 7.07M & 453.8M & 0.44J \\
        & 5 & 76.19 & 35.39M & 2269.32M & 2.21J \\
        \hline
        \multirow{2}{*}{CASCADE MT} & 1 & 74.80 & 7.07M & 784M & 0.74J \\
        & 5 & 76.59 & 35.39M & 3924.10M & 3.69J \\
        \bottomrule
    \end{tabular}
\end{table}




%% file: sections/conclusions.tex

\section{Discussions}

In this work, we propose Multiple Threshold (MT) approaches to mitigate the precision loss inherent in SNNs compared to floating-point neural networks. We demonstrate that MT-SNNs outperform state-of-the-art models with similar architectures, achieving higher accuracy with fewer time steps and comparable energy consumption, and raising the overall performance ceiling for SNNs. Crucially, the MT approach maintains compatibility with neuromorphic hardware by preserving the multiplication-free property of spiking neurons.



However, our work also reveals some limitations and potential avenues for future research:

\begin{itemize}[itemsep=2pt,topsep=0pt,parsep=0pt]
    \item The performance gains of MT on the CIFAR10-DVS dataset diminish at later time steps, potentially due to the limited size of this dataset. Exploring MT on larger-scale DVS datasets is an important direction for future work, although such datasets are currently scarce.
    \item The analogy between the CASCADE MT mode and activation quantization in ANNs suggests a potential bridge between these two types of neural networks. Investigating the conversion of pre-trained ANNs into CASCADE MT SNNs represents a compelling research direction, which could leverage the vast body of knowledge and pre-trained models available for ANNs to accelerate SNN development.
\end{itemize}

